\documentclass{article}
\usepackage{verbatim}

\usepackage{multirow} 
\usepackage{microtype}
\usepackage{graphicx}
\usepackage{subfigure}
\usepackage{booktabs} 
\usepackage{tabularx}
\usepackage{float}
\usepackage{hyperref}

\usepackage{float}  
\usepackage{booktabs}  
\usepackage{multirow}  

\usepackage[accepted]{icml2025}

\usepackage{amsmath}
\usepackage{amssymb}
\usepackage{mathtools}
\usepackage{amsthm}

\usepackage[capitalize,noabbrev]{cleveref}

\theoremstyle{plain}

\theoremstyle{definition}

\theoremstyle{remark}

\usepackage[textsize=tiny]{todonotes}
\usepackage{xcolor}

\usepackage[inline]{enumitem}

\icmltitlerunning{ Agentic Adversarial QA for Improving Domain-Specific LLMs}

\begin{document}

\twocolumn[
\icmltitle{ Agentic Adversarial QA for Improving Domain-Specific LLMs}



\icmlsetsymbol{equal}{*}

\begin{icmlauthorlist}
\icmlauthor{Vincent Grari}{axa,trail}
\icmlauthor{Ciprian Tomoiag\u{a}}{axa} 
\icmlauthor{Sylvain Lamprier}{leria}
\icmlauthor{Tatsunori Hashimoto}{stan}
\icmlauthor{Marcin Detyniecki}{axa,pol,trail}
\end{icmlauthorlist}

\icmlaffiliation{axa}{AXA Group Operations}
\icmlaffiliation{stan}{Stanford University}
\icmlaffiliation{leria}{LERIA, Universit\'e d’Angers, France}
\icmlaffiliation{trail}{TRAIL, Sorbonne Universit\'e, Paris, France}
\icmlaffiliation{pol}{Polish Academy of Science, IBS PAN, Warsaw, Poland}
\icmlcorrespondingauthor{Vincent Grari}{vincent.grari@axa.com}

\icmlkeywords{Machine Learning, ICML}

\vskip 0.3in
]



\printAffiliationsAndNotice{}  

\begin{abstract}

Large Language Models (LLMs), despite extensive pretraining on broad internet corpora, often
struggle to adapt effectively to specialized domains. 
There is growing interest in fine-tuning these models for such domains; however, progress is constrained by the scarcity and limited coverage of high-quality, task-relevant data. To address this, synthetic data generation methods such as paraphrasing or knowledge extraction are commonly applied. Although these approaches excel at factual recall and conceptual knowledge, they suffer from two critical shortcomings: (i) they  provide minimal support for interpretive reasoning capabilities in these specialized domains, and (ii) they often produce synthetic corpora that are excessively large and redundant, resulting in poor sample efficiency. 
To overcome these gaps, we propose an adversarial question-generation framework that produces a compact set of semantically challenging questions. These questions are constructed by comparing the outputs of the model to be adapted and a robust expert model grounded in reference documents, using an iterative, feedback-driven process designed to reveal and address comprehension gaps. Evaluation on specialized subsets of the LegalBench corpus demonstrates that our method achieves greater accuracy with substantially fewer synthetic samples.

\end{abstract}

\section{Introduction}
\label{submission}

Large Language Models (LLMs) pretrained on extensive internet-scale corpora have demonstrated remarkable general-purpose capabilities but often struggle to efficiently adapt to highly specialized, domain-specific knowledge contained within small, focused document sets. This difficulty arises primarily from the models’ unfamiliarity with critical domain-specific facts—which may appear only sparsely or even just once within a limited corpus—making it especially challenging to interpret and integrate these facts through complex reasoning. Although there is growing interest in fine-tuning LLMs for specific domains, progress has been limited due to the scarcity of specialized data, motivating the development of targeted synthetic augmentation approaches.

Recent approaches, such as EntiGraph \citep{yang2024entigraph} and Knowledge-Instruct \citep{ovadia2025knowledge}, aim to close the data-efficiency gap by generating synthetic training data from small corpora, typically through structured entity and fact extraction.  These methods commonly leverage a large, general-purpose model to generate new training data grounded in the specialized domain documents,  which is then used to improve a smaller, domain-specialized model, often via fine-tuning or distillation. While EntiGraph focuses on expanding entity-centric knowledge graphs for continued pretraining, Knowledge-Instruct reformulates extracted knowledge into instruction-response pairs for supervised fine-tuning. 
Although these approaches achieve strong performance on tasks focused primarily on factual recall, their effectiveness may diminish when faced with broader comprehension tasks, especially those requiring nuanced interpretation, inference, and integration of complex domain-specific knowledge.
For example, determining insurance coverage for an uncommon event, such as a client experiencing an accident while hiking abroad, requires interpreting rarely encountered or implicitly defined conditions within policy clauses. Resolving such questions involves not merely recalling explicit facts, but understanding nuanced language, subtle contextual implications, and layered dependencies within domain-specific documentation.

This observation suggests that it is not just the quantity of data that matters, but rather the quality and specificity of the augmentation strategies employed, particularly approaches explicitly designed to pinpoint and address specific model deficiencies. Existing augmentation methods generally generate synthetic data indiscriminately or based on pre-defined heuristics, without directly probing which concepts or reasoning tasks are most difficult for the model. This can lead to inefficient training, as much of the augmented data might reinforce knowledge the model already possesses. To address this, an adversarial approach naturally emerges as a promising alternative: by actively generating challenging questions tailored to expose precisely those aspects of reasoning and comprehension where the model exhibits weaknesses. 
In this manner, the model is exposed not only to questions that require factual recall but also, and more importantly, to those that demand interpretive and integrative reasoning. 
Inspired by principles from active learning~\cite{settles2009active}, boosting~\cite{freund1997decision}, and distributionally robust optimization (DRO)~\cite{duchi2021learning,sinha2017certifying}—which enhance learning by focusing on uncertain, misclassified, or worst-case instances where the model may perform poorly—we introduce an adversarial learning framework for smaller domain-specific LLMs. This framework iteratively generates questions, using feedback from a robust expert model, to systematically uncover and address interpretative weaknesses.

We demonstrate our adversarial feedback-driven methodology on specialized legal documents within the LegalBench corpus~\cite{guha2023legalbench}, showing how these targeted synthetic datasets substantially improve the domain-specific reasoning capabilities of smaller LLMs. Our method enables these models to achieve performance competitive with much larger counterparts, surpassing entity-centric and instruction-based strategies by more effectively addressing a wider range of nuanced comprehension challenges. 

\section{Related Work}
\textbf{Transfer Learning Across Domains.}
A core challenge in natural language processing is adapting language models to new domains where direct supervision or labeled data is scarce. Traditional approaches for cross-domain transfer learning often involve continued pretraining on domain corpora~\cite{gururangan2020don,devlin2019bert}, domain-adaptive fine-tuning~\cite{howard2018universal,lee2020biobert}, or leveraging multi-task learning frameworks~\cite{ruder2017overview}. Such methods aim to bridge the gap between general pretraining and the specific terminology, knowledge, or reasoning patterns required in target domains. However, their effectiveness may be constrained by the limited availability of high-quality domain-specific data, motivating the development of more data-efficient adaptation strategies.

\textbf{Model Distillation and Student-Teacher Paradigms.}
To further enhance transfer across domains, model distillation methods have been widely employed~\cite{hinton2015distilling,sanh2019distilbert}. In this setup, a large, general-purpose teacher model transfers knowledge to a smaller, specialized student model, typically by having the student mimic the outputs or intermediary representations of the teacher. Recent work has explored domain-adaptive distillation~\cite{turc2019well,jiao2019tinybert}, where the teacher is either fine-tuned or prompted for domain-specific tasks, and the resulting guidance is used to supervise smaller models with reduced resources. These teacher-student setups are related to our approach, but in our case, the teacher model is grounded in the same domain documents as the student. This ensures the feedback is tailored to the specific context, making the supervision more relevant and effective. 

\textbf{Synthetic Data Generation.}
Recent approaches adapt language models to specialized domains with limited data by generating synthetic corpora. Entity-centric methods such as EntiGraph~\cite{yang2024entigraph} expand on entities and relations via knowledge graphs, while paraphrase-based techniques~\cite{DBLP:conf/acl/MainiSBG0J24, ovadia2024} diversify training data by rewording existing texts. Instruction-driven approaches like Knowledge-Instruct~\cite{ovadia2025knowledge} convert extracted domain facts into instruction–response pairs, supporting efficient adaptation under low-resource settings. By contrast, our adversarial, feedback-driven framework improves sample efficiency and reasoning ability by selectively generating questions that require not only factual recall but also deeper context-dependent reasoning. 

Alongside these developments, there is increasing interest in using LLMs in iterative feedback loops to refine prompts and tasks. Such feedback-driven frameworks, where the model provides critiques or guidance, underlie our approach.

\textbf{Prompt Optimization via LLM Feedback.}
Recent approaches optimize prompts and inputs using automated feedback generated by large language models.
In these frameworks, prompt refinement are guided by structured critiques or optimization signals
from external LLMs. For instance, methods such as TextGrad~\cite{yuksekgonul2024textgrad} implement differentiable prompting, iteratively refining queries to maximize a model-evaluated reward. Other techniques leverage LLM feedback loops for instruction tuning and robust evaluation~\cite{yang2023large,madaan2023self}. Our adversarial
question generation pipeline builds upon this line of work, using LLM-driven feedback to provide fine-grained, diagnostic supervision in an adversarial optimization framework.

\section{Methodology}
Our objective is to enhance both the \textit{domain-specific knowledge}---the understanding and recall of concepts, facts, and language unique to a given specialized corpus---and the \textit{interpretive reasoning capabilities}---that is, the ability to accurately integrate, infer, and reason over such content---of language models, based solely on access to  
a domain document $C$ 
(e.g., a legal contract). As in prior work, we operate solely with the domain corpus itself, without requiring annotated datasets or downstream task supervision.

To systematically address both types of weaknesses, our approach introduces an iterative adversarial process that generates questions $Q^{(t)}$ targeting the model's factual gaps and interpretative limitations. For example, some questions may focus on domain-specific terminology, concepts, or facts explicitly present in $C$ (e.g., ``What is the definition of the term `subrogation' in this contract?''), while others may present challenging or hypothetical scenarios that require reasoning, inference, or integration of multiple clauses (e.g., ``If a client is injured while hiking abroad, would they be covered under this insurance policy?''). A model may, for instance, accurately define 'subrogation' but fail when asked to determine coverage in a scenario requiring integration of multiple clauses; an adversarial question targeting this scenario reveals such interpretive limitations. In this context, the effectiveness of adversarial question generation may be conceptually likened to pedagogical practices in active learning: rather than solely requiring the passive recall of explicitly stated definitions and isolated facts—analogous to students memorizing material for traditional written exams—comprehensive understanding is better gauged by posing challenging, integrated, or hypothetical scenarios akin to those one might encounter in rigorous oral examinations.

In our setting, we assume access to a large, highly capable language model which, when provided with the full domain corpus $C$ as context, can function as a robust “oracle” or expert. However, relying on such a system—in which the model always has access to all domain documents—is often impractical in real-world scenarios due to resource, latency, privacy, or deployment constraints. Therefore, our aim is to transfer domain expertise from this strong oracle model to a more lightweight, efficient target model.

\subsection{Adversarial Question Optimization}

We propose an adversarial optimization framework to systematically uncover and address interpretive deficiencies in domain-specific language models. This setup comprises two primary agents: a robust expert model ($f_{\text{strong}}$) and a target weaker model ($f_{\text{weak}}$), both provided with access to the same domain-specific context $C$. The core objective is to generate and iteratively refine questions about $C$ that maximize divergence between the responses of these models, thereby identifying aspects in which the target model exhibits limitations in either domain-specific knowledge or interpretive reasoning.
Formally, at each iteration $t$, we take the domain-specific context or document $C$ and the current question $Q_i^{(t)}$, 
and obtain answers from both models:
\[
A_{i, \text{strong}}^{(t)} = f_{\text{strong}}(C, Q_i^{(t)}), \quad
A_{i, \text{weak}}^{(t)} = f_{\text{weak}}(C, Q_i^{(t)}).
\]

We then evaluate the difference in their answers using a feedback function $f_{\text{fb}}$, which is typically instantiated as a capable LLM to compare the two responses and identify discrepancies along dimensions such as correctness, coverage, and contextual reasoning alignment (see Appendix~\ref{app:prompts}):
\[
\mathcal{L}(Q_i^{(t)}) = f_{\text{fb}}(A_{i, \text{strong}}^{(t)}, A_{i, \text{weak}}^{(t)}).
\]
The key objective of this adversarial process is to generate questions that maximize the measured disagreement, as quantified by $\mathcal{L}(Q)$, which serves as a proxy loss function indicating where the target model most strongly diverges from the expert's interpretive capacity. Specifically, we seek
\[
Q^{*} = \arg\max_{Q} \mathcal{L}(Q).
\]
To operationalize this iterative maximization over text, we adopt the differentiable prompting paradigm introduced in TextGrad~\cite{yuksekgonul2024textgrad}. Notably, while TextGrad is designed to minimize a task loss by refining prompts, our framework inverts this direction and explicitly maximizes interpretive disagreement in order to expose the limitations of the target model:
\[
Q_i^{(t+1)} = Q_i^{(t)} + \nabla_Q \mathcal{L}(Q_i^{(t)}).
\] 
Each refinement step comprises two stages: (i) a guidance model ($f_{\text{guide}}$) generates a natural language editing instruction, conditioned on the output of the feedback model ($f_{\text{fb}}$): $\nabla_Q \mathcal{L}(Q_i^{(t)}) = f_{\text{guide}}(Q_i^{(t)}, \mathcal{L}(Q_i^{(t)}))$. This instruction prescribes how to revise $Q_i^{(t)}$ to accentuate potential weaknesses and failure modes in the target model (see Appendix~\ref{app:prompts} for examples). (ii) a revision model ($f_{\text{rev}}$) applies this instruction to update the current question, yielding the next iteration, $Q_i^{(t+1)} = f_{\text{rev}}(Q_i^{(t)}, \nabla_Q \mathcal{L}(Q_i^{(t)}))$. 

In practice, we instantiate the three auxiliary agents---$f_{\text{fb}}$, $f_{\text{guide}}$, and $f_{\text{rev}}$---using the same strong LLM, each configured with a distinct instruction prompt (see Appendix~\ref{app:prompts}). This design preserves semantic consistency across modules while remaining implementation-lightweight. Notably, $f_{\text{guide}}$ and $f_{\text{rev}}$ correspond to the \texttt{TextGrad.backward} and \texttt{TextGrad.step} modules, respectively. 

Algorithm~\ref{alg:adversarial} summarizes the entire adversarial question optimization procedure, clearly delineating each step involved in refining questions to systematically identify and improve the interpretive limitations of the target model.

\subsection{Synthetic Dataset Construction and Fine-Tuning}

After completing the optimization procedure, we assemble a synthetic dataset consisting of the final adversarial questions paired with expert-provided answers:
\begin{equation*}
    \mathcal{D}_{\text{synthetic}} = \left\{ \left(Q_i^{(T)}, f_{\text{strong}}(C, Q_i^{(T)}) \right) \right\}_{i=1}^{N}.
\end{equation*}
Fine-tuning the weak model $f_{\text{weak}}$ on the expert-curated $\mathcal{D}_{\text{synthetic}}$ explicitly targets and remedies the shortcomings uncovered by adversarial optimization. Consequently, this focused training substantially enhances the model’s robustness and accuracy on domain-specific comprehension tasks.

\begin{algorithm}[H]
\caption{Iterative Adversarial Question Generation}
\label{alg:adversarial}
\begin{algorithmic}[1]
\STATE \textbf{Input:} Domain context $C$; initial question set $\{Q_i^{(0)}\}_{i=1}^N$; strong model $f_{\text{strong}}$; weak model $f_{\text{weak}}$; feedback model $f_{\text{fb}}$; guidance model $f_{\text{guide}}$; revision model $f_{\text{rev}}$; number of iterations $T$
\FOR{$i = 1$ {\bf to} $N$}
    \FOR{$t = 0$ {\bf to} $T-1$}
        \STATE $A_{i,\,\text{strong}}^{(t)} \gets f_{\text{strong}}(C, Q_i^{(t)})$
        \STATE $A_{i,\,\text{weak}}^{(t)} \gets f_{\text{weak}}(C, Q_i^{(t)})$
        \STATE $\mathcal{L}(Q_i^{(t)}) \gets f_{\text{fb}}(A_{i,\,\text{strong}}^{(t)},\, A_{i,\,\text{weak}}^{(t)})$
        \STATE $\nabla_Q \mathcal{L}(Q_i^{(t)}) \gets f_{\text{guide}}(Q_i^{(t)},\, \mathcal{L}(Q_i^{(t)}))$
        \STATE $Q_i^{(t+1)} \gets f_{\text{rev}}(Q_i^{(t)},\, \nabla_Q \mathcal{L}(Q_i^{(t)}))$
    \ENDFOR
\ENDFOR
\STATE \textbf{Return:} $\{Q_i^{(T)}\}_{i=1}^{N}$ 
\textit{(Final set of optimized 
questions)}
\end{algorithmic}
\end{algorithm}

\section{Experiments}

We assess our method on a targeted subset of the \textbf{LegalBench} benchmark~\cite{guha2023legalbench}, which originally comprises a wide range of tasks intended to evaluate the legal reasoning abilities of large language models. To emphasize domain-specific abilities, we focus on the three most frequently referenced contracts from the CUAD dataset~\cite{hendrycks2021cuad} within the LegalBench suite. These are: \textit{Cardlytics Maintenance Agreement}, \textit{Buffalo Wild Wings Franchise Agreement}, and \textit{PF Hospitality Franchise Agreement}. Across these contracts, there are a total of 491 benchmark questions spanning 36 distinct tasks. 

\begin{table}[!ht]
\centering
\caption{Accuracy (\%) across three contract-specific subsets and average for LLaMA3-8b.}
\label{tab:llama3-results}
\small
\begin{tabular}{lccccc}
\toprule
\textbf{Method} & Tokens & Cardl & Buffa & Pfhos & Avg \\
\midrule
\multicolumn{6}{l}{\textit{Baselines}} \\
No Extra Data & - & 67.3 & 69.1 & 72.1 & 69.5 \\
\midrule
\multicolumn{6}{l}{\textit{Ours}} \\
Ours & 96k & \textbf{82.7} & \textbf{79.6} & \textbf{85.7} & \textbf{82.7} \\
\midrule
\multicolumn{6}{l}{\textit{Competitor}} \\
Paraphrase $ \times 6 $ & 149k & 68.5 & 70.4 & 77.0 & 71.9 \\
Model-indep. QA & 147k & 75.0 & 74.1 & 78.3 & 75.8 \\
Entigraph & 6.7M & 80.4 & 76.5 & 82.0 & 79.6 \\
Knowledge\_Instr & 159k & 78.6 & 70.4 & 75.8 & 75.0 \\
\bottomrule
\end{tabular}
\end{table}

We compare our proposed fine-tuning approach against a base pretrained LLM , as well as  several increasingly sophisticated domain adaptation strategies. These include (1) paraphrase-based fine-tuning; (2) fine-tuning on uninformed questions, a naive synthetic dataset generated by prompting a strong LLM to write challenging questions for a given contract without any iterative refinement or adversarial objective; (3) EntiGraph-based augmentation \citep{yang2024entigraph}; and (4) Knowledge-Instruct \citep{ovadia2025knowledge}.
All models are evaluated using a few-shot setting with the standard LegalBench prompt template. To isolate the effects of fine-tuning strategies, we do not employ any retrieval augmentation or external tools.

Our adversarial QA generation approach consistently outperforms both paraphrase-based fine-tuning and model-independent questions fine-tuning baselines. It achieves a \textbf{18.99\%} improvement over the base model and a \textbf{3.89\%} gain relative to the EntiGraph competitor, while using $\approx$70$\times$ fewer training tokens. Whereas EntiGraph primarily augments entity-level relational knowledge, our feedback-driven question optimization specifically targets interpretive shortcomings, leading to richer semantic coverage and greater task accuracy. 

To assess the impact of iterative adversarial refinement, we perform a sensitivity analysis by varying the number of optimization steps during dataset construction. For each contract, we generate adversarial questions with different refinement iterations, fine-tune the target model, and evaluate on the corresponding LegalBench tasks.

\begin{figure}[!ht]
    \centering
    \includegraphics[width=1\linewidth]{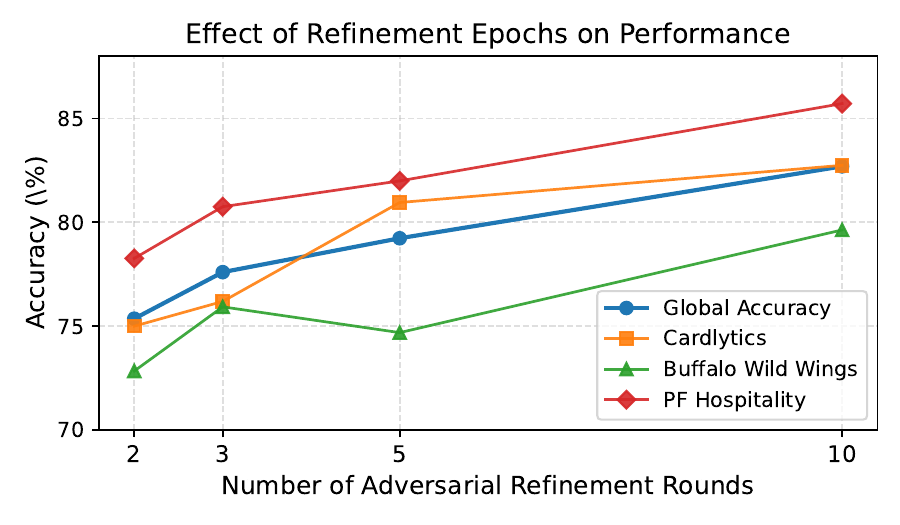}
    \caption{
        Accuracy on LegalBench (top three contracts) by epoch; 
        performance improves with more refinement.
    }
    \label{fig:ablation-epochs}
\end{figure}
As shown in Figure~\ref{fig:ablation-epochs}, additional refinement iterations result in progressively higher accuracy, highlighting the effectiveness of iterative adversarial feedback in improving domain-specific model performance.

\section{Conclusion}
We introduce an adversarial question generation framework to enhance domain-specific LLMs by systematically identifying and addressing their interpretive weaknesses. By adapting the TextGrad differentiable prompting method, our approach iteratively produces targeted questions and synthetic datasets aligned with model-specific shortcomings. Empirical results demonstrate that this strategy substantially improves the reasoning performance of smaller LLMs, even in limited data settings, highlighting the effectiveness of adversarial refinement for efficient domain adaptation.

\nocite{langley00}

\bibliography{example_paper}
\bibliographystyle{icml2025}

\clearpage

\appendix

\section{Appendix}
\subsection{Detailed Code Explanation}
This appendix provides an in-depth explanation of the implementation behind the adversarial question generation pipeline described in the main text. We outline its methodology, specific variable roles, code examples, and the adversarial optimization process.

\subsection{Overall Methodology}
The central goal of our approach is to iteratively generate adversarial questions that reveal and target the interpretive weaknesses of a smaller language model (termed the \emph{weak model}), using discrepancies with a larger, more robust expert model (the \emph{strong model}). The process relies on structured feedback from a feedback model, which evaluates the divergence between the models' responses in context.

\textbf{Algorithm explanation:}
\textbf{Algorithm explanation:}
\begin{itemize}
    \item \emph{Inputs:} A set of $N$ independently initialized questions (each starting from a placeholder, as described by the \texttt{question} variable in the next subsection), the fixed document context $C$, models $f_{\text{strong}}$, $f_{\text{weak}}$, and $f_{\text{fb}}$, as well as the total number of optimization iterations $T$.
    \item \emph{Response Generation:} For each question $Q_i$ at each iteration $t$, both models are queried with input $(C, Q_i^{(t)})$, yielding responses $A_{i, \text{strong}}^{(t)}$ and $A_{i, \text{weak}}^{(t)}$.
    \item \emph{Discrepancy Evaluation:} The feedback model $f_{\text{fb}}$ compares these responses, computing both a numeric disagreement score and a natural language explanation, denoted $\mathcal{L}(Q_i^{(t)})$.
    \item \emph{Gradient Computation:} The guidance model $f_{\text{guide}}$ receives the explanation from the feedback model and generates a natural language edit instruction, indicating how to revise $Q_i^{(t)}$ so as to further increase the answer discrepancy.
    \item \emph{Question Update:} The revision model $f_{\text{rev}}$ uses the current question and the edit instruction to generate an updated question, thereby steering the question to maximize divergence between the models’ outputs.

\end{itemize}
After $T$ rounds, the final set of optimized questions $\{Q_i^{(T)}\}_{i=1}^N$ together with their strong model answers are saved for constructing the synthetic dataset.

\medskip

\noindent\textbf{Implementation Note:} \\
In practice, the variable \texttt{question} is initialized as a learnable object (`requires\_grad=True`) and is updated via feedback-driven optimization. All model weights and the context document remain fixed; only the content of the question is iteratively revised.

\subsection{Implementation Details}
Our implementation uses the \texttt{textgrad} framework, leveraging a differentiable prompting paradigm and modern LLM APIs. Below, each major component and step is described.

\subsubsection{Model Configuration}
Three key model interfaces are initialized via \texttt{LiteLLMEngine}:
\begin{itemize}
\item \textbf{Strong Model}: Provides authoritative answers strictly based on the contract. In practice, we use gpt-4o-mini.
\item \textbf{Weak Model}: The target of improvement, typically smaller or less domain-specialized. In practice, we use llama3.1:8b.
\item \textbf{Feedback Model}: Evaluates discrepancies between strong and weak model answers to guide question optimization.
\end{itemize}

\subsubsection{Central Role of the Optimized \texttt{question} Variable}

A distinctive and central aspect of our method is that the \texttt{question} variable itself is the object of optimization. Instead of tuning model parameters, our pipeline holds both the contract text and all model weights fixed---and iteratively refines the natural language content of the \texttt{question} prompt.

\textbf{In code, the variable is initialized as:}
\begin{quote}
\texttt{question = tg.Variable(}\\
\texttt{~~"Q: ???",~~\# minimal placeholder}\\
\texttt{~~requires\_grad=True,}\\
\texttt{~~role\_description="A legal question to be optimized for maximal response divergence"}\\
\texttt{)}
\end{quote}

\textbf{Conceptually,} at each iteration, the current value of \texttt{question} is updated to maximize measured discrepancy between the weak and strong models. This iterative process adaptively generates natural language questions that are adversarial: they are optimized to be especially likely to expose errors, misunderstanding, or blind spots in the weak model, given a fixed contract context.

\subsubsection{Prompt Templates}
\label{app:prompts}
Carefully crafted prompts ensure the models fulfill their roles.

\textbf{Strong and Weak Models Prompt:}
\begin{quote}
\texttt{You are an expert in interpreting contracts.}\\
\texttt{The user will provide a contract and a question about it.}\\
\texttt{Provide the most accurate, thorough answer based on the contract's text.}\\
\texttt{Stay strictly within the contract's details and do not invent external laws.}
\end{quote}

\textbf{Feedback Model Prompt:}
\begin{quote}
\texttt{You are an expert in legal contract analysis.}\\
\texttt{Given two responses (correct vs.\ potentially incorrect), identify contradictions, omissions, or errors in the second response.}\\
\texttt{Provide a numeric incorrectness score (0.0 to 1.0) with detailed explanation.}
\end{quote}

\textbf{Example formatted prompt for discrepancy evaluation:}
\begin{quote}
\texttt{Compare the following responses:}\\
\texttt{\textless CONTRACT\textgreater: \{contract\}}\\
\texttt{\textless QUESTION\textgreater: \{question\}}\\
\texttt{\textless CORRECT\_RESPONSE\textgreater: \{response\_strong\}}\\
\texttt{\textless POSSIBLY\_INCORRECT\_RESPONSE\textgreater: \{response\_weak\}}\\
\texttt{Rate the incorrectness of the second response (0 to 1) and explain errors, contradictions, or missing details.}
\end{quote}

\subsubsection{Optimization Algorithm}
Optimization employs the TextGrad framework's differentiable prompting technique. Each iteration refines \texttt{question} based on textual feedback, with code logic as follows:

\begin{quote}
\texttt{optimizer.zero\_grad()}\\
\texttt{divergence\_loss = loss\_fn(contract\_text, question, response\_strong, response\_weak)}\\
\texttt{divergence\_loss.backward()}\\
\texttt{optimizer.step()}
\end{quote}

This iterative update is \emph{conceptually} analogous to a gradient ascent step,
since each feedback-driven edit aims to increase (rather than decrease) the disagreement proxy---but the updates are performed purely in the space of natural language.

\subsection{Synthetic Dataset Construction}
After optimization, synthetic adversarial questions and their authoritative answers (from the strong model) are saved to a dataset. This dataset is constructed specifically to target and address the interpretative weaknesses surfaced in the weak model.

\subsection{Illustrative Example: Iterative Adversarial Refinement of Legal Questions}

To concretely illustrate our feedback-driven methodology, we present an example drawn from adversarial QA generation over a complex legal contract—the “Software License, Customization and Maintenance Agreement” between Bank of America and Cardlytics, Inc. At each iteration $t$, our algorithm interacts with the document context $C$, refines the current question $Q^{(t)}$, and leverages feedback from model disagreements to systematically probe where the target model falls short.

\paragraph{Initial Setup}
\begin{itemize}
    \item \textbf{Document Context ($C$):} Raw, multi-page legal contract (e.g., 30+ pages covering IP, confidentiality, audit, etc.).
    \item \textbf{Current Question $Q^{(t)}$:} The natural-language prompt being optimized.
    \item \textbf{Strong (Expert) Model:} Large LLM (e.g., GPT-4o-mini in our experiments) with robust reasoning over $C$.
    \item \textbf{Weak (Target) Model:} Smaller LLM (e.g.,  Llama-3 8B in our experiments); less accurate on nuanced legal reasoning.
    \item \textbf{Feedback Model:} Compares strong/weak responses, assigns a numerical error score, and provides actionable natural-language critique highlighting omissions, misconceptions, or interpretive errors.
\end{itemize}

\paragraph{Iteration 0: Initialization}

\textbf{Q$^{(0)}$:} \texttt{Q: ???}

\textbf{Strong Model’s Response:}
\emph{“It appears that you did not provide a specific question regarding the contract. Please clarify your question, and I will do my best to provide an accurate and thorough answer based on the contract’s text.”}

\textbf{Weak Model’s Response:}
\emph{“…it appears that you are asking about a specific aspect… If I were to guess, one possible question could be: 'What are the requirements for background checks…'…”}

\textbf{Feedback Model Output:}
\begin{itemize}
    \item \textbf{Score:} 0.5 (partially incorrect/misleading)
    \item \textbf{Critique:} Weak model guesses rather than prompting for clarification; fails to require user input.
    \item \textbf{TextGrad Guidance:} “Replace vague placeholder with a meaningful, contextually grounded question targeting a specific clause (e.g., confidentiality, indemnity). Use hypotheticals or contrast with industry standards.”
\end{itemize}

\textbf{Revision $\rightarrow$ Q$^{(1)}$:}
\texttt{What are the potential risks and benefits associated with the confidentiality provisions in this contract, and how might these affect both parties in terms of liability and operational flexibility?}


\paragraph{Iteration 1: Sharpening Scope and Depth}

\textbf{Strong Model:} Multi-paragraph answer, citing specific sections, discussing operational and compliance implications.

\textbf{Weak Model:} Discusses generic risks/benefits but omits contract-specific definitions and scenario detail.

\textbf{Feedback Model Output:}
\begin{itemize}
    \item \textbf{Score:} 0.6 (incomplete)
    \item \textbf{Critique:} Omits key details (e.g., definition of “confidential info”, obligations on breach, liability nuances).
    \item \textbf{TextGrad Guidance:} “Expand comparative scope—request scenario analysis, stakeholder view, and cross-industry contrast (e.g., healthcare, GDPR).”
\end{itemize}

\textbf{Revision $\rightarrow$ Q$^{(2)}$:}
\texttt{Considering the entire Agreement, what are the potential risks and benefits associated with the confidentiality provisions, and how do these provisions compare to those in other financial services contracts? Can you provide examples or scenario analyses for their impact on liability and operational flexibility for both parties and stakeholders involved?}


\paragraph{Iteration 2: Multi-Perspective and Hypothetical Analysis}

\textbf{Strong Model:} Expands with cross-industry contrasts (e.g., with HIPAA), stakeholder-specific discussion, hypothetical adverse event scenarios.

\textbf{Weak Model:} Improves, but still superficial on scenario depth and long-term effects.

\textbf{Feedback Model Output:}
\begin{itemize}
    \item \textbf{Score:} 0.6
    \item \textbf{Critique:} Insufficient detail in long-term consequences, stakeholder-specific risk perception.
    \item \textbf{TextGrad Guidance:} “Ask for best- vs. worst-case hypotheticals, speculative future impact, and divergent interpretations by different actors.”
\end{itemize}

\textbf{Revision $\rightarrow$ Q$^{(3)}$:}
\texttt{How do the confidentiality provisions, liability clauses, and compliance requirements in this agreement influence operational flexibility and stakeholder perspectives? How do they compare to industry standards across healthcare, technology, and privacy? Provide contrasting hypothetical scenarios and discuss possible long-term consequences of breaches or compliance failures (including differing stakeholder interpretations).}


\paragraph{Final feedback at $T$ iteration} 

This adversarial, iterative loop yields a maximally challenging question, such as:

\texttt{In what ways do the confidentiality provisions, liability clauses, and compliance requirements influence operational flexibility and stakeholder perspectives? How do these elements compare with industry standards across sectors such as healthcare and technology? Can you provide contrasting hypothetical scenarios showing potential risks and benefits, with a future-facing discussion on how evolving laws and technologies might force adaptation? What differing interpretations might various stakeholders bring to these elements?}

\end{document}